\documentclass[letterpaper, 10 pt, conference]{ieeeconf}  

\IEEEoverridecommandlockouts                              

\usepackage{graphics} 
\usepackage{epsfig} 
\usepackage{mathptmx} 
\usepackage{times} 
\usepackage{amsmath} 
\usepackage{amssymb}  
\usepackage{algorithm,algorithmic}
\usepackage{amsfonts}

\usepackage{subfigure}
\usepackage{graphicx}
\usepackage{epstopdf}
\usepackage{booktabs}
\usepackage{multirow}
\usepackage{cite}

\usepackage{threeparttable}
\usepackage{tabularx}
\newcolumntype{Y}{>{\centering\arraybackslash}X}
\newcolumntype{Z}{>{\raggedleft\arraybackslash}X}

\title{\LARGE \bf
Learning Residual Model of Model Predictive Control via Random Forests for Autonomous Driving
}

\author{Kang Zhao$^{1}$, Jianru Xue$^{1}$, Xiangning Meng$^{1}$, Gengxin Li$^{1}$, and Mengsen Wu$^{1}$  
\thanks{*This work was not supported by any organization}
\thanks{$^{1}$All authors are with Institute of Artificial Intelligence and Robotics, College of Artificial Intelligence, Xi'an Jiaotong University, Xi'an, China. Corresponding author's email:
        {\tt\small jrxue@mail.xjtu.edu.cn}}%
\thanks{}%
}

\begin{document}

\maketitle
\thispagestyle{empty}
\pagestyle{empty}

\begin{abstract}

One major issue in learning-based model predictive control (MPC) for autonomous driving is the contradiction between the system model's prediction accuracy and computation efficiency. 
The more situations a system model covers, the more complex it is, along with highly nonlinear and nonconvex properties. These issues make the optimization too complicated to solve and render real-time control impractical.
To address these issues, we propose a hierarchical learning residual model which leverages random forests and linear regression.
The learned model consists of two levels. The low level uses linear regression to fit the residues, and the high level uses random forests to switch different linear models. 
Meanwhile, we adopt the linear dynamic bicycle model with error states as the nominal model.
The switched linear regression model is added to the nominal model to form the system model. 
It reformulates the learning-based MPC as a quadratic program (QP) problem, and optimization solvers can effectively solve it. 
Experimental path tracking results show that the driving vehicle's prediction accuracy and tracking accuracy are significantly improved compared with the nominal MPC.
Compared with the state-of-the-art Gaussian process-based nonlinear model predictive control (GP-NMPC), our method gets better performance on tracking accuracy while maintaining a lower computation consumption. 

\end{abstract}

\section{Introduction}

 Model predictive control (MPC)\cite{camacho2013model} has been a new trend in the application of autonomous driving and racing. However, it suffers from the problems of modeling difficulty and solving efficiency, as it relies on a sufficiently descriptive model of the system to optimize performance and ensure constraint satisfaction, which renders modeling critical for the success of the resulting control system\cite{hewing2020learning}. 
 However, the traditional physics-based models make many assumptions to simplify the modeling process, which leads to large errors during mid and long-term predictions. 
 To overcome the modeling problems, many researchers employ machine learning methods to learn the system models and improve the control performance. 
 There are two mainstream modeling methods using machine learning: the data-driven and hybrid models. 
 
 The data-driven model learns entirely from the input-output data collected from the system and does not use any physical information from the system model.
 It suits the situation when there is no physical information about the system, or it is too hard to build the system model. 
 However, it relies on a large number of training data and lacks generalization to cases out of the training data, which may result in collisions on autonomous driving applications. 
 On the contrary, the hybrid model only learns the unknown part of the system model based on the physical model. It has the advantage of interpretability, achieves better prediction performance, and relies on fewer data.
 A typical method of hybrid modeling is to learn the residual model of the system \cite{kabzan2019learning}.
 The residual model means the difference between the actual system output and the nominal model.

 The learning methods of the residual model can be divided into probabilistic and deterministic ones. 
 Probabilistic models, such as
 Bayesian linear regression (BLR)\cite{mckinnon2019learning} and Gaussian process (GP)\cite{kabzan2019learning}, are typical methods of learning the residual model. 
 However, the distribution estimation via the Bayesian model is typically not tractable unless in the case of a linear dynamic model with Gaussian noise and Gaussian prior of the parameter distribution\cite{hewing2020learning}. 
 Gaussian process-based nonlinear model predictive control (GP-NMPC) \cite{kabzan2019learning}, a masterpiece of learning-based methods for automatic racing applications, takes the nonlinear dynamic bicycle model as the nominal model. Therefore the optimization has to use methods like the sequential quadratic program (SQP) to take the GP residues into the objective function iteratively. As a result, the GP with a large size of training data makes the optimization intractable. So it can't use the full potential of the collected data and may lead to large errors. On the contrary, deterministic models, such as feedforward neural networks, convolutional neural networks, and long short-term memory networks. have the ability to learn from large volumes of data. However, the above deterministic models using deep learning techniques will make efficiency problems as the learned models are highly nonlinear and complex, which makes the control-oriented optimization intractable.

\begin{figure*}[t]
    \scriptsize
		\includegraphics[width=0.96\textwidth]{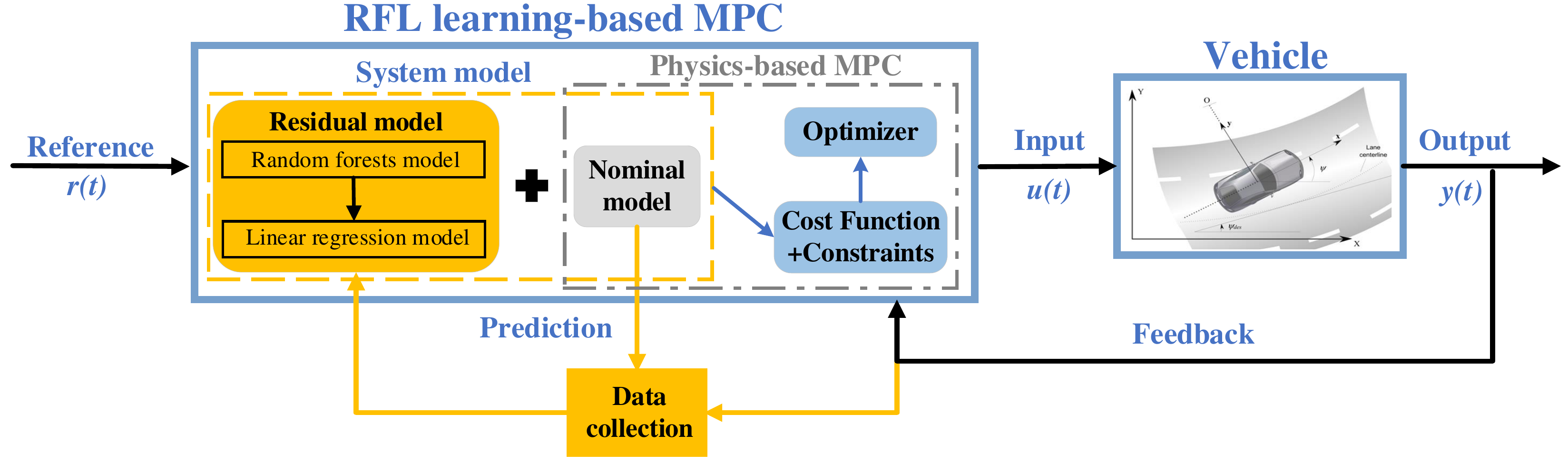}
		\\
	\caption{System diagram of random forests linear (RFL) learning-based model predictive controller (MPC). The typical physics-based MPC diagram uses the nominal model as the system model, shown in the gray dotted line box, and the RFL-MPC uses the residual model added to the nominal model as the system model. The residual model uses random forests and linear regression to learn the residue collected from the history data running by nominal MPC.} 
	\label{system}
\end{figure*}

To address the aforementioned issues, we propose a hierarchical learning residual model via random forests and linear regression. 
The work has the following contributions: (1) We propose a hybrid model in which the nominal model is a dynamic bicycle model with error states, and the residual model is a hierarchical learning model. 
Combining the nominal model with the learned model improves the prediction performance of the predictive model so as to improve the tracking accuracy.
(2) The low layer of the hierarchical residual model uses multiple linear regressions to fit the residue, and the higher layer uses random forests to switch different linear models. 
The additive model of the system is formed by the addition of the linear regression models and the linear dynamical bicycle model with error states. 
It makes the path tracking a quadratic program (QP) problem and reduces the solution complexity greatly. There are several efficient optimization solvers for QP problems, which enable the learning-based MPC to obtain real-time control performance. Meanwhile, the time consumption of prediction by random forests and linear regression is independent of the number of training data, which indicates a potential to cover all the states of the system.


\section{Related Works}


Data-driven models of vehicles\cite{spielberg2019neural,da2020modelling} can be divided into system identification and end-to-end model learning. 
System identification\cite{vicente2020linear} employs the subspace system identification method to estimate the equation parameter matrix of a linear time-invariant (LTI) state-space model.
The data-driven LTI model is of higher accuracy than the physics-based linear and nonlinear model with estimated parameters, but it still makes large errors during long-term prediction as it loses the nonlinear part of the dynamics.
To obtain highly accurate approximation, a two-hidden layer neural network\cite{spielberg2019neural} and Long Short-Term Memory (LSTM) \cite{xu2019automated} are adopted to learn the vehicle dynamics. Simulation\cite{xu2019automated} results show that LSTM surpasses the physics-based and MLP models in accuracy and time cost.
However, the above data-driven models are only used by simple control algorithms or simulations. 
As the learned models are highly nonlinear and complex, they are not suitable for multi-step prediction control with optimization, such as MPC. 
Identifying an accurate vehicle model that is suitable for control-oriented methods only using a data-driven model is still challenging due to the high nonlinearity. 

The residual models of autonomous vehicles can be divided into probabilistic and deterministic ones. Probabilistic models, such as
 BLR\cite{mckinnon2019learning} and GP\cite{kabzan2019learning} are typical methods of learning the residual model. Hewing et al. \cite{hewing2018cautious} use GP to learn the uncertainties and mismatch of the dynamic model and extend the NMPC problem with the learned additive model in a stochastic way. However, the computation time is the main challenge for a naive implementation of GP-NMPC, as the computation complexity depends on the number of data points used. To address this problem, computation approximations( e.g., inducing points\cite{quinonero2005unifying} and selected basis functions\cite{lazaro2010sparse}), local GPs\cite{meier2016towards}, sparse GPs\cite{hewing2020learning}, data selection, and fixed size dictionary of used points are proposed. Based on the techniques above, many applications of robotic systems using GP-NMPC get great success, e.g., path following of mobile robots\cite{ostafew2016learning,ostafew2016robust}, trajectory tracking of robotic manipulation\cite{carron2019data}, and autonomous racing\cite{kabzan2019learning}. Especially, J. Kabzan et al. \cite{kabzan2019learning} implement the GP-NMPC on FSD vehicles and prove the hybrid model control can achieve much better performance on tracking accuracy and safety than the physics-based dynamic model. However, as the computation complexity of Gaussian process regression relies on the measured sample data, the learning data has to maintain a volume of 300 data points to guarantee the calculation time, which limits the application range to short distances like repetitive racing cars and restricts the tracking accuracy to some extent.

Weighted Bayesian linear regression (wBLR)\cite{mckinnon2019learning} is adopted to learn a changing part of the actuator dynamics and test it on a ground vehicle with a speed of less than 2m/s. Experiments on the muddy test track show that the wBLR has better prediction accuracy and fewer tracking errors than the GP with 50 data points. However, the experiments are held on a short distance and repetitive path tracking tasks, which can't prove the feasibility of long-term and high-speed path tracking. 

Deterministic models have the ability to learn from large volumes of data. However, to the best of our knowledge, there are rarely deterministic models used in the learning of residual models, as the learned models are highly nonlinear and complex, which makes the control-oriented optimization intractable. On the contrary, the method that we proposed has the potential to overcome the above problems. 

In summary, the learning models of vehicle dynamics are rarely used in the MPC of real autonomous vehicles due to their highly nonlinear and nonconvex properties. Instead, they are mainly on repetitive tasks and control-in-the-loop simulations. In order to overcome the shortcomings of the existing learning methods, we propose the hierarchical learning method as follows.

\section{Overview of RFL-MPC}
The core demand of hybrid modeling in MPC is that the data-driven model has the ability to cover most cases and can't bring too much computation complexity to the optimization solver. To leverage sufficient description and computation efficiency, an intuitive thought is to learn switching residual models for different situations. By switching different residual models, especially linear residual models, during the optimization, the computation complexity will be greatly decreased. Inspired by the ideas above,  we propose to use random forests and linear regression to learn the residual models. The corresponding control methods are called random forests linear learning-based model predictive control (RFL-MPC), and the system diagram is shown in Fig. \ref{system}.

 In detail, the process can be divided into offline learning and online control two stages. In the offline learning stage, we use the typical physics-based MPC to collect data. The physics-based MPC uses only a nominal model $\mathit{f_{nom}}$  (section IV-B) to predict the future system states and optimize the control variables to get desired performance, as shown in the gray dotted line box. However, there are mismatches (between the real system and the nominal model) and environmental uncertainties. Here, we use the residual model to represent the difference between the real system $\mathit{f_{true}}$ and the nominal model $\mathit{f_{nom}}$ at the time step of $\mathit{k+1}$. Then a Random Forests Linear learning (section V-B) method learns the residue part as the residual model. The random forests learn the residue as a rough model and switch the samples to different leaves. And then, the linear regression model learns from samples on each leaf as a fine model. In the online control stage, the learned residual model is added to the nominal model to reform the system model as the additive model (section IV-A). The online control uses the additive model to predict the future system states and optimize the control variables to get better performance on accuracy and safety.

\section{Modeling for Vehicle Control}
\begin{figure}[t]
	\begin{center}
		\includegraphics[width=0.42\textwidth]{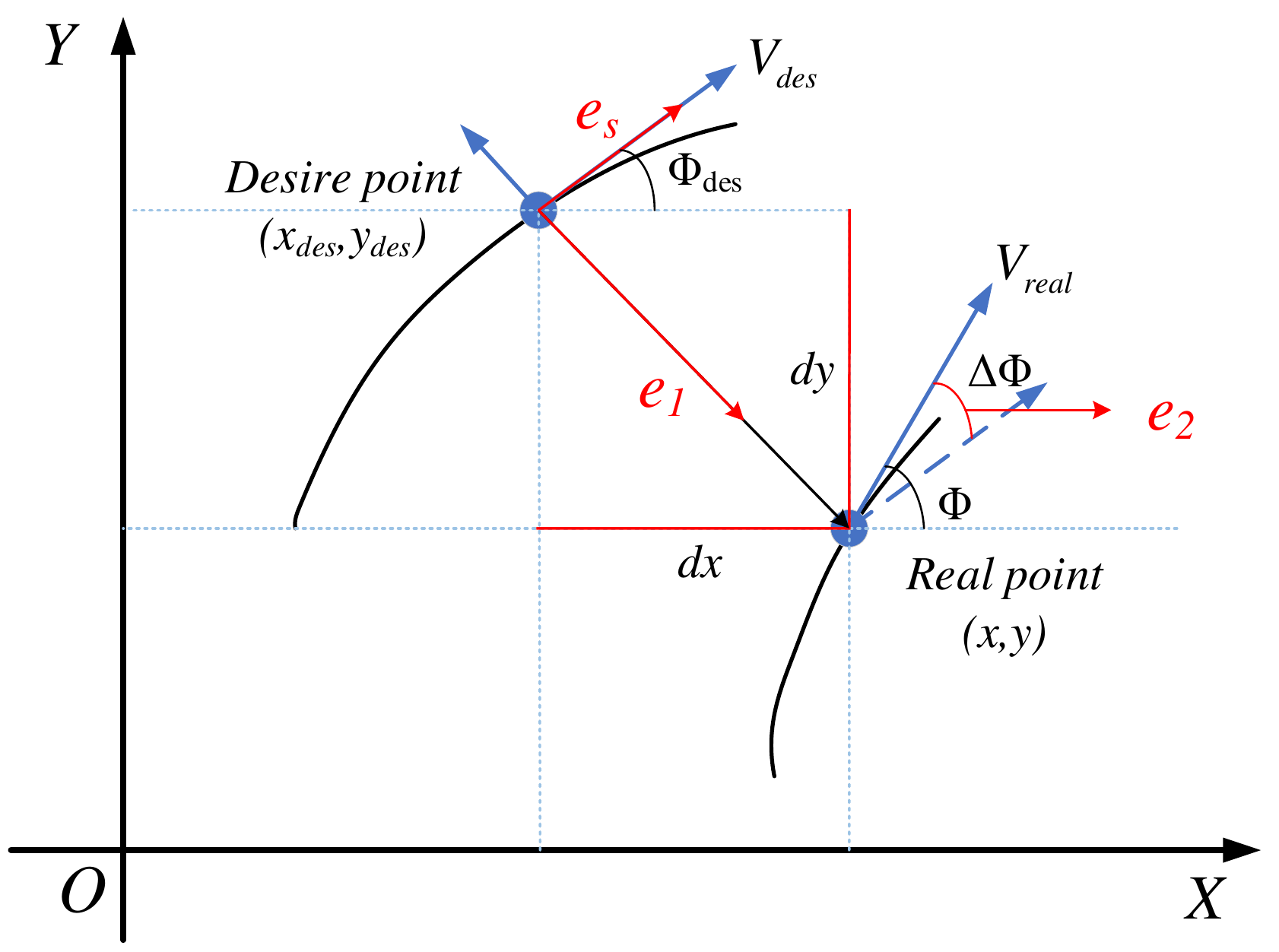}
	\end{center}
	\caption{
		Dynamic bicycle model with error states.
        }
	\label{modeling}
\end{figure}

Modeling is the most essential part of MPC, and there are several models for autonomous driving, such as the kinematic model, dynamic model, and dynamic model with error states. The dynamic bicycle model with error states is one of the most commonly used models for autonomous vehicle control since it takes both the simplicity for onboard calculation and high accuracy for taking performance control into consideration. The dynamic bicycle model with error states is shown in Fig. \ref{modeling}, with control variables $\mathit{u=[\delta]}$ and states
\[
	\textbf{x}=[e_{1};\ \dot{e}_{1};\ e_{2};\ \dot{e}_{2}]
\]
where $\mathit{e_{1},e_{2}}$ are the lateral position error and yaw angle error of the vehicle with respect to the desired path, $\mathit{\dot{e_{1}},\dot{e_{2}}}$ are the corresponding first-order derivatives, and $\delta$ is the steering angle of the front wheel. 

\subsection{Additive Model}

The considered  additive model is of the form
\begin{equation}
    \textbf{x}_{k+1}=f(\textbf{x}_{k},u_{k})+r(\textbf{z}_{k})
\end{equation}
where $\mathit{f(\cdot)}$ means the nominal vehicle model (section IV-B) and $\mathit{r(\cdot)}$ means the residual model, the difference between the real states of the vehicle and the predictive states of the nominal vehicle model.

\subsection{Nominal Vehicle Model}

The nominal vehicle model assumes the car with a rigid body and linear tire forces, traveling with constant longitudinal velocity, on a road of constant radius $\mathit{R}$, and can be expressed as 
\begin{equation}
\dot{\textbf{x}}=\begin{bmatrix}\dot{y}+v_{x}(\phi-\phi_{des})
 \\(\ddot{y}+v_{x}\dot{\phi})-v_{x}^{2}/R
 \\\dot{\phi}-\dot{\phi}_{des}
 \\\ddot{\phi}-\ddot{\phi}_{des}
\end{bmatrix}
\end{equation}
where $\mathit{\dot{y},\ddot{y}}$ represent the vehicle lateral velocity and acceleration, $\mathit{\phi,\dot{\phi},\ddot{\phi}}$ represent the vehicle yaw angle, velocity, and acceleration, the subscript $des$ means the desired ones, and $\mathit{v_{x}}$ represents the longitudinal vehicle velocity.

To constrain the step variation of control variables, we build the increment model by adding $u_{k-1}$ to the states, and thus the increment states can be expressed by $\textbf{x}_{k}=[e_{1, k};\dot{e}_{1, k};e_{2, k};\dot{e}_{2, k};u_{k-1}]$. The control variable is expressed by $\Delta u_{k}=u_{k}-u_{k-1}$.

\subsection{Evolution Model}

Utilizing linearization and discretization \cite{falcone2007nonlinear}, the above nominal vehicle model under the future $N$ time steps can take the shape of
\begin{equation}
\begin{aligned}
\textbf{X}_{k}&=\Psi_{k}\xi_{k}+\Phi_{k}\Delta U_{k}+\gamma_{k}
\\\eta_{k}&=\textbf{C}\textbf{X}_{k}
\end{aligned}
\end{equation}
where $\textbf{X}_{k}$, $\eta_{k}$, $\Delta U_{k}$ are the future states, outputs, and control variables of the next $N$ steps, respectively. $\xi_{k}$ is the matrix of initial states.
\[
\begin{aligned}
\textbf{X}_{k}&=\begin{bmatrix}\textbf{x}_{k+1|k};\  \textbf{x}_{k+2|k};\ \cdots;\ \textbf{x}_{k+N|k} \end{bmatrix}
\\\eta_{k}&=\begin{bmatrix}\textbf{y}_{k+1|k};\  \textbf{y}_{k+2|k};\ \cdots;\ \textbf{y}_{k+N|k} \end{bmatrix}
\\\Delta U_{k}&=\begin{bmatrix}\Delta u_{k|k};\ \Delta u_{k+1|k};\ \cdots;\ \Delta u_{k+N-1|k} \end{bmatrix}
\\\xi_{k}&=\begin{bmatrix}\textbf{x}_{k|k};\ \textbf{x}_{k|k};\  \cdots;\ \textbf{x}_{k|k} \end{bmatrix}
\end{aligned}
\]
where the subscript $i|k$ means the $i$-th step prediction at the $k$-th time step.
Coefficient matrices  $\gamma_{k}$, $\Psi_{k}$, $\Phi_{k}$ are as follows
\[
\begin{aligned}
\\\Psi_{k}=\begin{bmatrix}A_{k}
 \\A^2_{k}
 \\\vdots 
 \\A^N_{k}
 \end{bmatrix}, \
\gamma_{k}=\begin{bmatrix}D_{k}
 \\A_{k}D_{k}+B_{k}
 \\\vdots 
 \\\sum_{i=0}^{N-1}A^i_{k}D_{k}
\end{bmatrix}, \
\textbf{C}=\begin{bmatrix}C_{k}
 \\C_{k}
 \\\vdots 
 \\C_{k}
 \end{bmatrix}
\end{aligned}
\]
\[
\begin{aligned}
\Phi_{k}=\begin{bmatrix}&B_{k}
 \\&A_{k}B_{k} &B_{k}
 \\&\vdots &\vdots &\ddots
 \\&A^{N-1}_{k}B_{k} &A^{N-2}_{k}B_{k} &\cdots &B_{k}
\end{bmatrix}
 \end{aligned}
\]
where matrices $\mathit{A}_{k}$, $B_{k}$, $C_{k}$, $D_{k}$ are the linearized and discretized system matrix of (2) at time step $k$, and for details, see references \cite{falcone2007nonlinear,rajamani2011vehicle}.

\section{Learning Residual Model}

The central idea of learning residual models is using machine learning tools to obtain the control-oriented models that the MPC or other optimal control methods can still be used to solve the optimization problem feasibly and efficiently. Generally speaking, the residual model can be considered as a black-box model $\mathit{r(\textbf{z}_{k})=r(\textbf{x}_{k},u_{k},\textbf{d}_{k})}$, where $\mathit{\textbf{x},u,\textbf{d}}$ represent the states, inputs, and disturbances, respectively. Depending upon the system and the nominal model, $\mathit{r(\cdot)}$ is basically nonlinear, nonconvex, and nondifferentiable. Thus, the learning representations which make the optimization problem intractable can’t be directly applied to the optimization and control. As a result, we decompose the complex problem into two parts: (1) nonlinear, nonconvex, and nondifferentiable part learning by Regression Trees (RT) or Random Forests (RF) and (2) linear part learning by least-squares fitting.
\begin{equation}
r(\textbf{x}_{k},u_{k},\textbf{d}_{k})=I(g(\textbf{x}_{k},\textbf{d}_{k})==j)h_{j}(u_{k})
\end{equation}
where the learning part $\mathit{h(\cdot)}$ is linear, convex, and differentiable, with control variables $\mathit{u_{k}}$, and $\mathit{g(\cdot)}$ is somewhat nonlinear, nonconvex, and nondifferentiable with state variables and disturbances. $\mathit{I(\cdot)}$ is the characteristic function which equals 1 when $\mathit{g(\cdot)}$  puts the sample to the $j$-th leaf and activates the $j$-th linear function $\mathit{h_{j}(\cdot)}$. The linear decomposition or the variable separation can make up for the disadvantages of the complex residual model.

\subsection{Learning Residue with Regression Trees}
An intuitive thought of regressing nonlinear functions is to divide the data space into small pieces where a small number of samples fit by simple models. As a result, we divide the residual model Eq.(4) into two parts, the tree regression part $\mathit{g(\cdot)}$ and the linear approximation model $\mathit{h(\cdot)}$ for each leaf on the tree.

\begin{figure}[t]
	\begin{center}
		\includegraphics[width=0.48\textwidth]{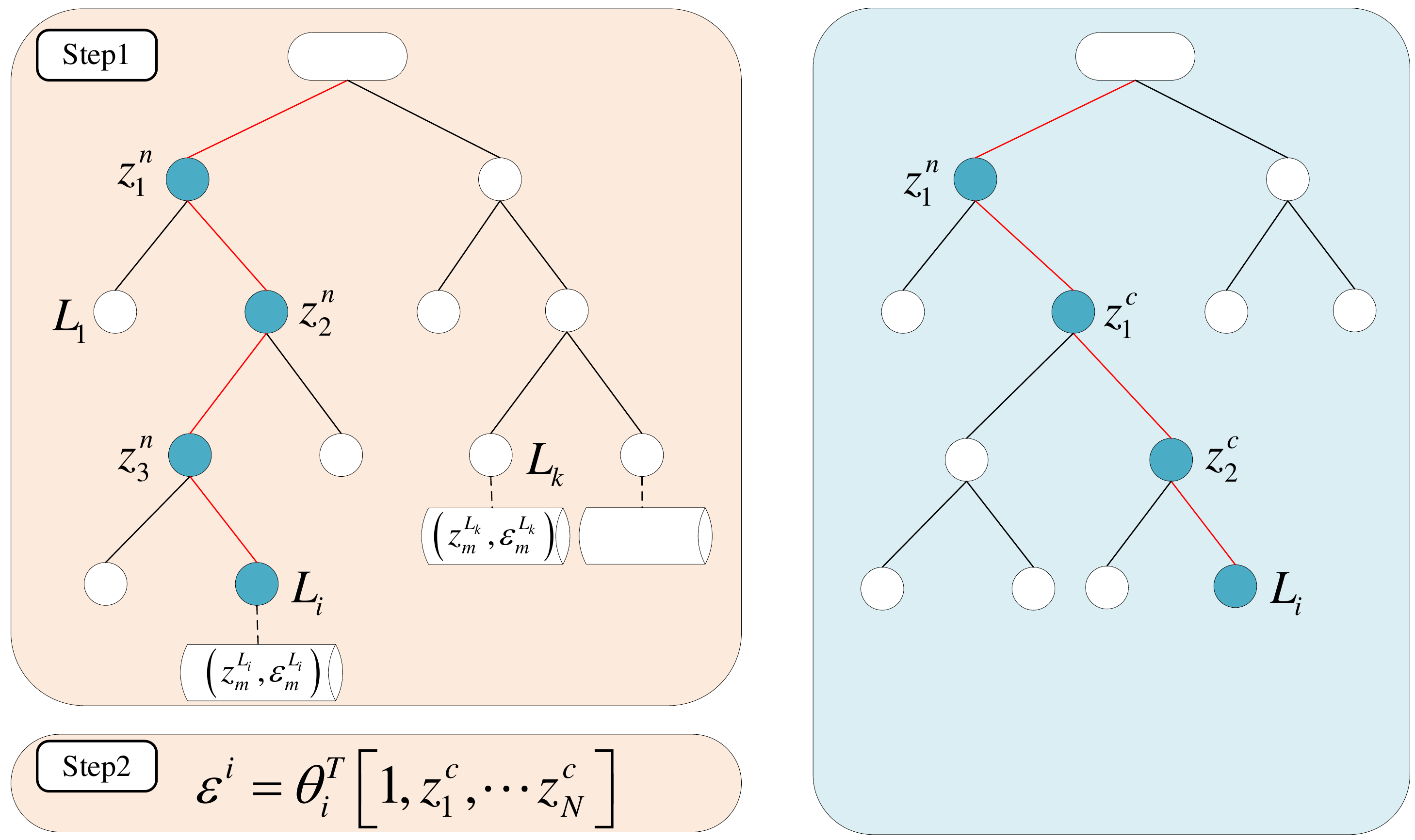}
	\end{center}
	\caption{
		Separation of variables. Step 1: Tree $\mathit{T_{1}}$ is trained on the features of non-control variables $\mathit{\textbf{z}^{n}}$ and every leaf $\mathit{L_{i}}$ is the output of the tree $\mathit{T_{1}}$ with $\mathit{m}$ samples on each leaf. Step 2: In the leaf $\mathit{L_{i}}$, a linear regression model is trained with the samples on each leaf as the function of the control variables $\mathit{\textbf{z}^{c}}$. Tree $\mathit{T_{2}}$ uses both control variables and non-control variables to regress the residue and is difficult to compute for control.}
	\label{train}
\end{figure}

The goal now is to predict the future residue $\mathit{\epsilon}$ at current time $k$ for the next $N$ steps, i.e., $\mathit{\epsilon_{k+1|k},\epsilon_{k+2|k},\cdots,\epsilon_{k+N|k}}$, where the $\mathit{N}$ represents the prediction horizon. Here, we divide the related variables into two sets: control/manipulated variables $\mathit{\textbf{z}^{c}\in {\mathbb{R}}^{c}}$ and non-control variables $\mathit{\textbf{z}^{n}\in {\mathbb{R}}^{n}}$. And the two sets compose the full feature set, i.e., $\mathit{\textbf{z}\equiv \textbf{z}^{c}\cup \textbf{z}^{n}\in {\mathbb{R}}^{c+n}}$. There are two main reasons for the separation of the variables\cite{jain2017data}: (1) The objective functions generally consist of state variables and control variables, especially for QP problems; (2) Control variables of the future steps are unknown and need to be optimized. Our goal is to use the linear approximation to represent the residual model of the system, so we consider the residual variables $\mathit{\epsilon\in {\mathbb{R}}^{r}}$ as the outputs for training, where $\mathit{\epsilon}$ can either represent $r$ regression trees with single output or one regression tree with $r$ outputs. The training data is in the form of $\mathit{(\textbf{z}_{i},\epsilon_{i})}$, and the training process is composed of two steps, as shown in Fig. \ref{train}.

Using the non-control variables, we build regression trees by CART \cite{breiman2017classification} and the system's $j$-th predictive residue $\mathit{\epsilon_{k+j|k}}$ depends on the states before the $j$-th state.
\begin{equation}
\begin{aligned}
\epsilon_{k+j|k}=f_{tree}(\textbf{z}^n_{k+j-N|k},\cdots,\textbf{z}^n_{k+j-1|k})
\\ \textbf{z}^n_{k+j-1|k} \in {\mathbb{R}}^{n},\forall j=1,2,\cdots,N
\end{aligned}
\end{equation}
After the first step, the training data is divided onto different leaves, and on each leaf, $\mathit{L_{i}}$, the linear model is defined as
\begin{equation}
\begin{aligned}
\epsilon_{k+j|k}=\theta^T_{i}[1,\textbf{z}^c_{k+j-N|k},\cdots,\textbf{z}^c_{k+j-1|k}]
\\ \textbf{z}^c_{k+j-1|k} \in {\mathbb{R}}^{c},\forall j=1,2,\cdots,N
\end{aligned}
\end{equation}
As the samples on each leaf are different, so linear coefficients $\mathit{\theta_{i}}$  on each leaf are different. Just as equation (6) shows, to get a suitable form for the QP solver, the prediction residue $\mathit{\epsilon_{k+j|k}}$ is the linear combination of the control variables from time $\mathit{k+j-N}$ to time $\mathit{k+j-1}$. 

\subsection{Learning Residue with Random Forests}

There are drawbacks to using regression trees to predict the residue and divide the samples into the leaves since a single tree is accompanied by high variance and easily overfits the data. Small changes in data may lead to different feature selections and splits and thus affect the prediction accuracy and the division of the samples. To overcome the high variance and overfit problems, we combine several independent regression trees using ensemble methods \cite{hastie2009elements}. The central idea in ensemble methods is to reduce the overall variance by averaging noisy trees. There are two ways to get randomness: (1) using bootstrapped/sub-sampled training data and (2) randomly picking the features/variables to build the tree. And thus, trees in the forest will be trained with different training data and split into different nodes and leaves.

With random forests, we take the place of (5) by
\begin{equation}
\begin{aligned}
\epsilon_{k+j|k}=f_{forest}(\textbf{z}^n_{k+j-N|k},\cdots,\textbf{z}^n_{k+j-1|k})
\\ \textbf{z}^n_{k+j-1|k} \in {\mathbb{R}}^{n} ,\quad \forall j=1,2,\cdots,N
\end{aligned}
\end{equation}
where the forests train on a non-control variable set $\textbf{z}_{n}$, and the full variables samples are divided into different leaves of different trees. And then, we fit a linear regression model by control variables $\textbf{z}_{c}$ on every leaf $\mathit{L_{l,i}}$ on every tree $\mathit{T_{l}}$. 
\begin{equation}
\begin{aligned}
\epsilon_{k+j|k}=\Theta^T_{l,i}[1,\textbf{z}^c_{k+j-N|k},\cdots,\textbf{z}^c_{k+j-1|k}]
\\ \textbf{z}^c_{k+j-1|k} \in {\mathbb{R}}^{c},\quad \forall j=1,2,\cdots,N,\forall l=1,2,\cdots,M
\end{aligned}
\end{equation}

The random forests will improve the prediction accuracy, lower variance, and get a better division of samples on leaves at the cost of training time and a slight time burden in run-time. A forest of $M$ trees has $M$ linear coefficients, and a simple way to obtain a unified linear coefficient $\Theta_{j}$ is to average the $M$ linear coefficients.

\subsection{Learning MPC with Random Forests}

The above training process is done offline. But there are problems when predicting the future residues in run-time, as the $\mathit{k+j}$ time step prediction residue $\mathit{\epsilon_{k+j|k}}$ needs the system states from time $\mathit{k+j-N}$ to time $\mathit{k+j-1}$, i.e., $\mathit{\textbf{z}^n_{k+j-N|k},\cdots,\textbf{z}^n_{k+j-1|k}}$, while we only get the states previous to the current time, i.e., $\mathit{\textbf{z}^n_{k+j-N|k},\cdots,\textbf{z}^n_{k|k}}$. As a technique, we take the control solution of the last time step $\mathit{u_{k-1|k-1},u_{k|k-1},\cdots,u_{k+N_{c}-1|k-1}}$ as the initial control of this time step and assume the last variable keeps still. Thus the initial control $\mathit{u_{k|k-1},\cdots,u_{k+N_{c}-1|k-1},u_{k+N_{c}-1|k-1}}$ can be used to estimate the future $N$ step prediction states at the current time step with the nominal evolution model (3). Then in run-time, given the previous states $\mathit{\textbf{z}^n_{k+j-N|k},\cdots,\textbf{z}^n_{k|k}}$ and the last control, we can get all the states needed to narrow down to a leaf in (7) to apply the linear models in (8).
With the linear models in (8), the nominal evolution equation can be replaced by the full evolution equation.
\begin{equation}
\begin{aligned}
\textbf{X}_{k}&=\widehat{\Psi}_{k}\xi_{k}+\widehat{\Phi}_{k}\Delta U_{k}+\widehat{\gamma}_{k}
\\\eta_{k}&=\textbf{C}\textbf{X}_{k}
\end{aligned}
\end{equation}
where new matrix coefficient $\mathit{\widehat{\psi}_{k}=\psi_{k}}$, $\widehat{\Phi}_{k}=\Phi_{k}+\Delta \Phi_{k}$, $\widehat{\gamma}_{k}=\gamma_{k}+\Delta \gamma_{k}$, and
\[
\begin{aligned}
\\\Delta\widehat{\Phi}_{k}=\begin{bmatrix}&\beta_{1,N}
 \\&\beta_{2,N-1} &\beta_{2,N}
 \\&\vdots &\vdots &\ddots
 \\&\beta_{N,1} &\cdots &\beta_{N,N-1}  &\beta_{N,N}
\end{bmatrix}
\end{aligned}
\]
\[
\begin{aligned}
\Delta\widehat{\gamma}_{k}&=\begin{bmatrix}\beta_{1,0}+\sum_{i=1}^{N-1}\beta_{1,i}u_{k+i-N|k}
 \\\beta_{2,0}+\sum_{i=1}^{N-2}\beta_{2,i}u_{k+i+1-N|k}
 \\\vdots 
 \\\beta_{N,0}+\sum_{i=1}^{1}\beta_{N,i}u_{k-1|k}
\end{bmatrix}
\end{aligned}
\]
where $\beta_{j,n}$ means the $n$-th coefficient in the unified coefficients vector $\Theta_{j}$ of the $(k+j)$-th prediction at time step $k$.
The corresponding MPC problem can be formulated as
\begin{equation}
\begin{aligned}
\arg\min_{u_{k:k+N-1}} \parallel \eta -\eta ^{ref} \parallel_{Q_{1}} +\parallel \bigtriangleup U \parallel_{Q_{2}}+\lambda \sigma  
\end{aligned}
\end{equation}
\[
\begin{aligned}
s.t.\textbf{X}_{k}&=\widehat{\Psi}_{k}\xi_{k}+\widehat{\Phi}_{k}\Delta U_{k}+\widehat{\gamma}_{k}
\\\eta_{k}&=\textbf{C}\textbf{X}_{k}
\\\underline{\Delta U} &\le \Delta U\le \overline{\Delta U } 
\\\underline{\eta} &\le \eta\le \overline{\eta } 
\end{aligned}
\]
where $\eta ^{ref}$ represents the reference; $Q_{1}$ and $Q_{2}$ represent the weight coefficients of states and control variables; $\underline{\Delta U},\overline{\Delta U },\underline{\eta}, and \overline{\eta }$ represent the lower limit and the upper limit of control variables and the states. The slack variables $\sigma$ ensure recursive feasibility by relaxing the states, and $\lambda$ is the corresponding weight. We solve the above equation to get the optimal control sequence $[u_{k|k},...u_{k+N-1|k} ]$ by QP and only apply the $\mathit{1^{st}}$ control variable $u_{k|k}$ to the system. And resolve the above problem by the next time step.
The training and run-time procedure are given in Algorithm 1.

\begin{algorithm}  
	\caption{Model Predictive Control with Random Forests}  
	\label{alg1}
	\begin{small}  
	\hspace*{0.02in}Training Process (Off-line) \\
	\hspace*{0.02in}{\bf Procedure:} Model Training with RF and Linear approximation\\
	\vspace{-10pt}
	\begin{algorithmic}[1] 
    	\STATE {Input Set $\mathit{\textbf{z}^c\in {\mathbb{R}}^{c}} \leftarrow$ control \  variables }
		\STATE{Input Set $\mathit{\textbf{z}^n\in {\mathbb{R}}^{n}} \leftarrow$ non-control \  variables}
		\STATE{ Build Random Forests using training data $\mathit{(\textbf{z}^n,\epsilon)}$ with CART algorithm, get $T$ trees and each tree has $M_{i}$ leaves. Each leaf $\mathit{L_{i,j}}$ consists of samples $\mathit{(\textbf{z}^{n+c},\epsilon)^{L_{i,j}}}$}
        \FOR {tree from {$\mathit{L_{1}}$} to {$\mathit{L_{T}}$}}
		\FOR {leaf from {$\mathit{L_{i,1}}$} to {$\mathit{L_{i,M_{i}}}$}}
			\STATE{Linear regression to get $\Theta_{i,j}$ using LS}
		\ENDFOR\\
        \ENDFOR\\
	\leftline{\bf End procedure} 
	\leftline{Run Time} 
	\leftline{{\bf Procedure:}Model Predictive Control with Residue} 
		\WHILE{$\mathit{k<k_{stop}}$}
		\STATE{Estimate the predicted states of the next $N$ steps  by equation (3) and initial control input variables $\mathit{u_{k|k-1},\cdots,u_{k+N_{c}-1|k-1},u_{k+N_{c}-1|k-1}}$}\\
		\STATE{Using the random forests to narrow down the estimated states of the next $N$ steps to $N$ leaves on each tree}
		\FOR{estimated states from $\mathit{\eta^{est}_{1}}$ to $\mathit{\eta^{est}_{N}}$}
		\STATE{Decide the corresponding leaf $\mathit{L_{i,j}}$ as in (7)}
		\STATE{Get the linear coefficients $\mathit{\theta_{i,j}}$ on $\mathit{L_{i,j}}$ as in (8)}
        \STATE{Averaging $\Theta_{i,j}$ to get $\Theta_{j}$}
		\ENDFOR
		\STATE{Add the linear coefficients $\mathit{\beta_{i,j}}$ to the new evolution equation (9). Solve the optimal problem of (10) with the control sequence $\mathit{u_{k|k},\cdots,u_{k+N-1|k}}$, and apply the $\mathit{1^{st}}$ input $\mathit{u_{k|k}}$.}
		\STATE{k++}
		\ENDWHILE\\
	\leftline{\bf End procedure} 
	\end{algorithmic}
\end{small}
\end{algorithm}

\subsection{Computing Complexity Analysis}

The computing complexity of different learning methods is listed in TABLE \ref{table1}. As GP-NMPC\cite{kabzan2019learning} uses the nonlinear bicycle dynamic model and learns with sparse GP, the optimal solver, e.g., SQP has to query the model several times at each timestep. The predictive mean complexity is around  $O(Iter*m)$, and predictive variance complexity is around $O(Iter*m^{2})$, where $Iter$ means the max optimization iterations set artificially, and $m<n$ as the inducing points is a small set selected from the training data. The method that we proposed uses the linear dynamic model and switched linear learning approximation, which can be solved by QP solvers such as OSQP\cite{stellato2020osqp}, qpOASES\cite{ferreau2014qpoases}, and  HPIPM\cite{frison2020hpipm}. Without iteratively taking the learning part into optimization iterations and not being affected by the training data size, the predictive mean complexity is around  $O(FD)$ or $O(TFD)$. In other words, the predictive mean complexity of RFL is $O(1)$ considering the size of training data, which means it has more potential to harness large amounts of training data and save time compared with the GP method. 

\begin{table}[h]
\caption{Computing Complexity Comparison with Different Residual Models}
\label{table_example}
\begin{center}
\begin{tabularx}{0.48\textwidth}{l c c c c}
\toprule
&GP	&Sparse GP	&RT	&RF	\\
\midrule
Train time 	&$O(n^3)$	&$O(m^2n)$	&$O(nFD)$	&$O(T*\tau\log_{}{\tau})$\\

Train space 	&$O(n^2)$	&$O(mn)$	&$-$	&$-$\\

Predict mean 	&$O(n)$	&$O(m)$	&$O(FD)$	&$O(T*FD)$	\\
Predict variance  &$O(n^2)$	&$O(m^2)$	&$-$	&$-$\\
\bottomrule
\end{tabularx}
\end{center}
\begin{tablenotes}
\footnotesize
\item[]{Where $n$ is the sample number of the training data, $m$ is the number of inducing points. The complexity data of GP and Sparse GP are mainly from\cite{lazaro2010sparse,liu2020gaussian}. For RT and RF, $F$ is the feature number, $D$ is the tree depth, $T$ is the number of trees in the forest, and $\tau=nFD$.}
\end{tablenotes}
\label{table1}
\end{table}

\section{Experiments}

\begin{figure}[t]
    \centering
    \scriptsize
           
		\includegraphics[width=0.42\textwidth]{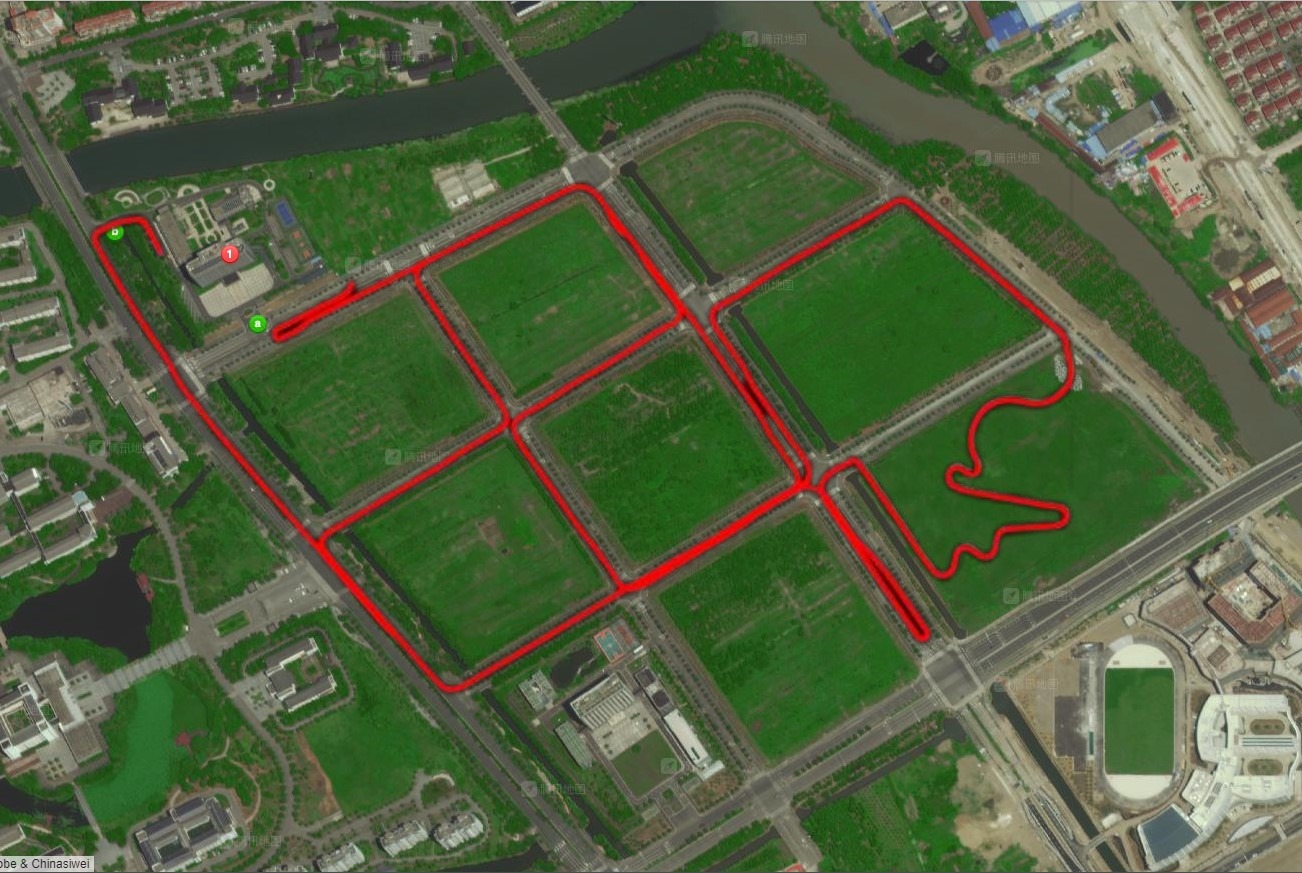} 
            
	\caption{Changshu Sudoku map. The red lines show the training reference used in the experiments.}
	\label{map}
\end{figure}

The path-following experiments are tested on the joint simulation of Matlab/Simulink and Carsim. The complex nonlinear car model comes from Carsim with a calculated rate of 1000Hz, and the path-following algorithms, such as MPC come from the Matlab/Simulink platform. The reference path comes from the Changshu Sudoku, shown in Fig. \ref{map}, the competition area of the Intelligent Vehicle Future Challenge (IVFC) in China. The reference consists of different kinds of straight lines, right/left turns, U-turns, and cross-country roads.

\subsection{Data Collection and Offline Learning}

The training data comes from the MPC with only the nominal model, with a prediction horizon of $N_{p}=16$ and a control horizon of $N_{c}=16$ at a sampling time of $T_{s}=0.02s$. The training input consists of the non-control variables of vehicle states (i.e., $e_{1},e_{2},\dot{e_{1}},\dot{e_{2}}$) and the control variables. The corresponding training labels are the residual part calculated by (1). The measuring indices are Root Mean Square Error (RMSE) and Max Error (ME) on both training and test data. An index of Mean Absolute Error (MAE) will be used to measure the tracking quantity in section VI-B. The RMSE, ME, and MAE are defined as
\begin{equation}
\begin{aligned}
RMSE&=\sqrt{\frac{1}{l}\sum_{k=1}^{l}\left ( \textbf{y}_{k}-\hat{\textbf{y}} _{k}  \right )   }
\\ME&=\max |\left ( \textbf{y}_{k}-\hat{\textbf{y}} _{k}\right |,k=1,\dots ,l  
\\MAE&= \frac{1}{l}\sum_{k=1}^{l}|\left ( \textbf{y}_{k}-\hat{\textbf{y}} _{k}  \right ) |
\end{aligned}
\end{equation}
where $\textbf{y}$ and $\hat{\textbf{y}}$ are the true output and the predicted value, and $l$ is the number of samples.

\begin{figure}[t]
	\centering
	\subfigure[Residue prediction with RT/RF and linear regression]{
		\begin{minipage}[b]{0.45\textwidth}
			\includegraphics[width=1\textwidth]{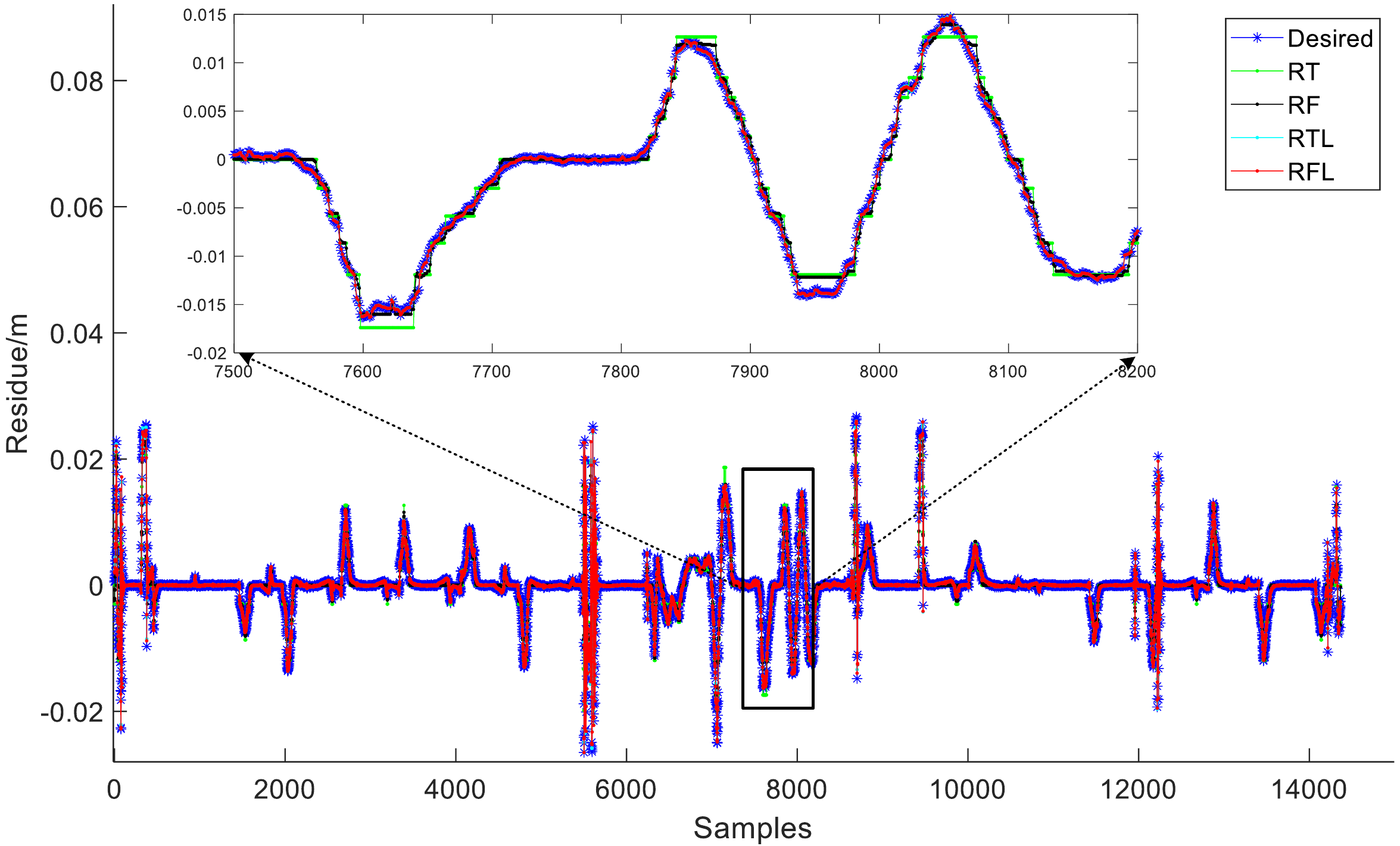} 
		\end{minipage}
        \label{case1}
	}
	\subfigure[Fitting error with RT/RF and linear regression]{
		\begin{minipage}[b]{0.45\textwidth}
			\includegraphics[width=1\textwidth]{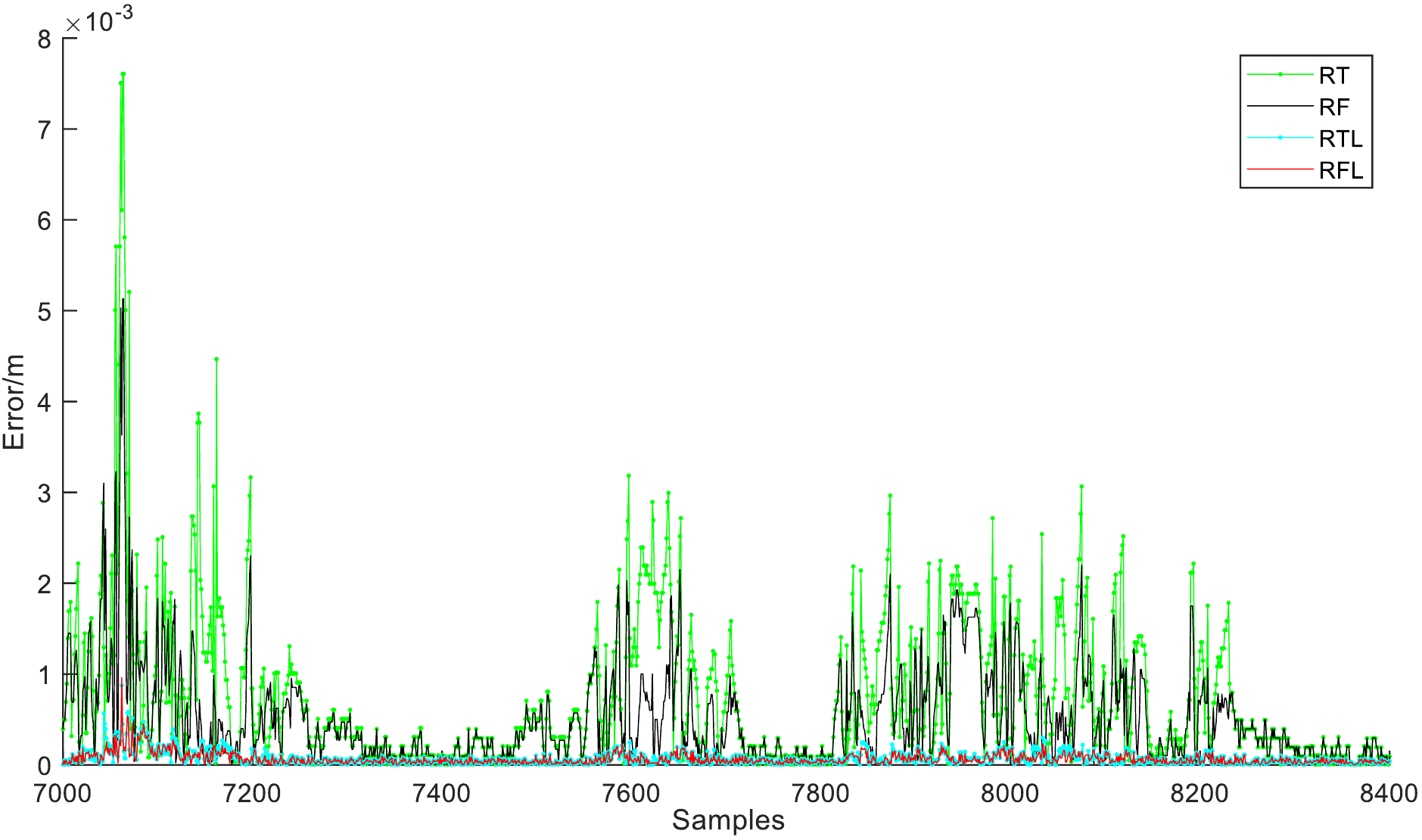}
		\end{minipage}
        \label{case2}
	}
	\caption{Residue prediction and fitting error with regression trees/ random forests and linear regression.}
	\label{result1}
\end{figure}

The regression trees were trained with the input of non-control variables and output labels and split by the CART algorithm. The leaves show the prediction of residue and consist of the corresponding samples.
The linear regression trains with training input of control variables and training output label on each leaf by the least square method. The prediction with RT and linear regression is shown in Fig. \ref{result1}, and the fitting error lies in TABLE \ref{table2}, and it shows that with the linear regression (RTL) the fitting of residue is greatly improved by RMSE from 1.57‰ to 0.1479‰, and ME from 3.19cm to 0.34cm.

The training data of random forests goes in line with the regression trees, while the features/variables and the trained samples are picked randomly to get randomness. The time-consuming shows a positive correlation with the tree numbers, to keep the regression and optimization in time, here we chose tree number 20. As shown in Fig. \ref{result1} and TABLE \ref{table2}, the fitting results are slightly improved by RF than RT (by RMSE from 1.57‰ to 1.48‰), while the linear regression part gets 10 times less fitting error than that without it (by RMSE from 1.48‰ to 0.1477‰). And the random forests linear regression (RFL) gets the best prediction on the test data of both RMSE and ME.
\begin{table}[h]
\caption{Fitting RMSE of Different Regression Methods on Test Data}
\label{table_example}
\begin{center}
\begin{tabularx}{0.48\textwidth}{l Y Y Y Y Y}
\toprule
&Data &	RT&	RF&	RTL	&RFL\\
\midrule
\multirow{2}*{RMSE\ (‰)}	&Training&	1.60&	1.52&	0.158&\textbf{0.156} \\
&	Test&	1.57 	&1.48 &	0.1479 	&\textbf{0.1477}\\
\midrule
\multirow{2}*{ME\ (cm)}&	Training&	2.72&	2.80&	0.33&	\textbf{0.31}\\
	&Test	&3.19	&2.81	&0.34	&\textbf{0.32}\\
\bottomrule
\end{tabularx}
\end{center}
\label{table2}
\end{table}

\subsection{Online Control with RT/RF and Linear Regression}

The online control with learning residual models using RT/RF and linear regression are tested on some test references out of the Sudoku. From TABLE \ref{table3}, we can see that the MPC+RTL gets a lot of improvement on the lateral tracking MAE by 14.29\% and The  MPC+RFL obtains the most improvement on the MAE by 16.32\%, while the RT and RF without linear regression obtain only 7.89\% and 9.02\% improvement. There are some drawbacks to adopting the residual model and RF method. From TABLE \ref{table4}, we can see that the average computing time of MPC+RTL and MPC+RFL is greatly increased as a result of dividing the future states into the leaves, and with more trees in the forest, the computing time will increase linearly. It is a trade-off between accuracy and efficiency. 

\begin{figure}[t]
	\centering
	\subfigure[Right turn of test path tracking]{
		\begin{minipage}[b]{0.45\textwidth}
			\includegraphics[width=1\textwidth]{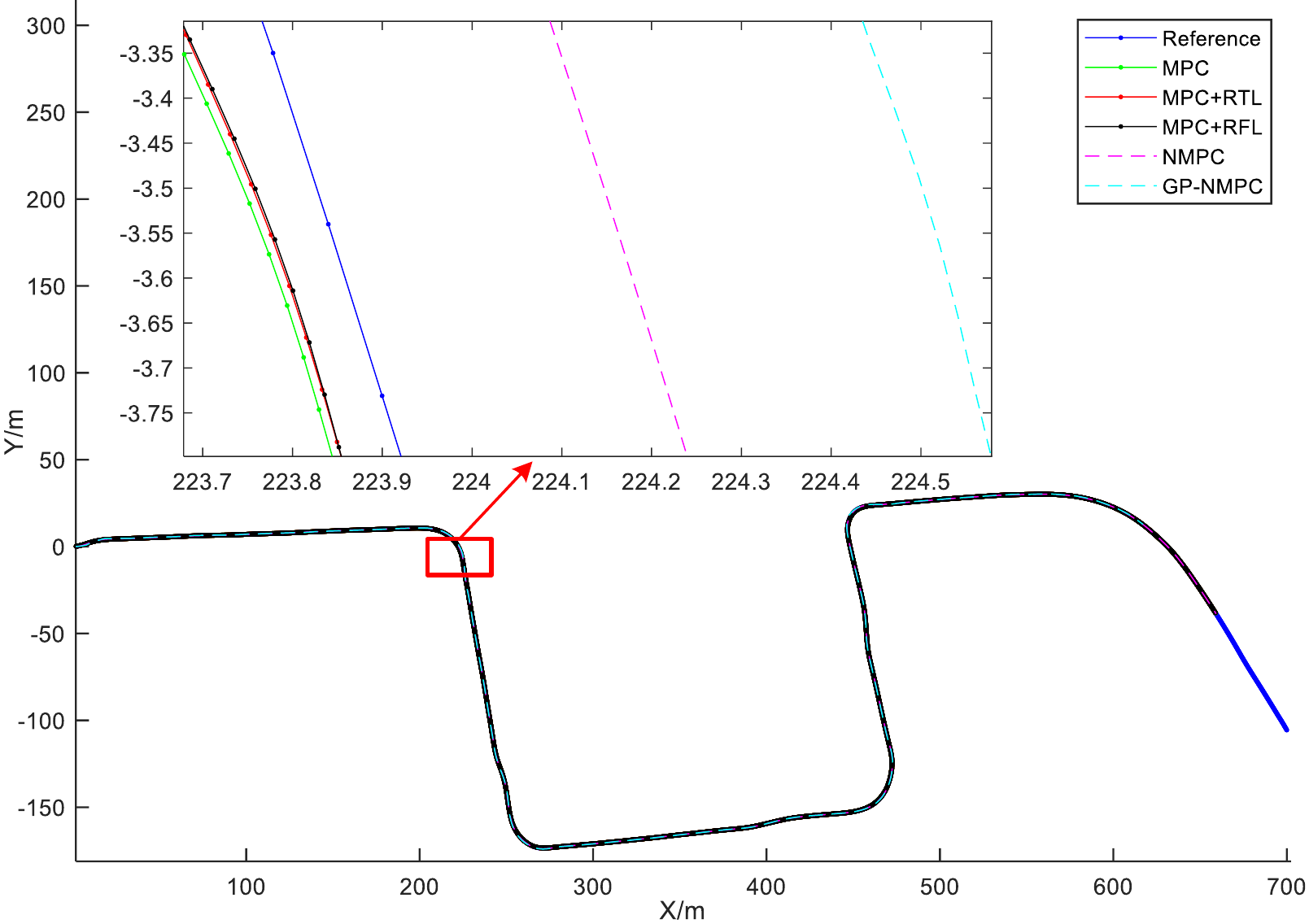} 
		\end{minipage}
        \label{case1}
	}
	\subfigure[Left turn of test path tracking]{
		\begin{minipage}[b]{0.45\textwidth}
			\includegraphics[width=1\textwidth]{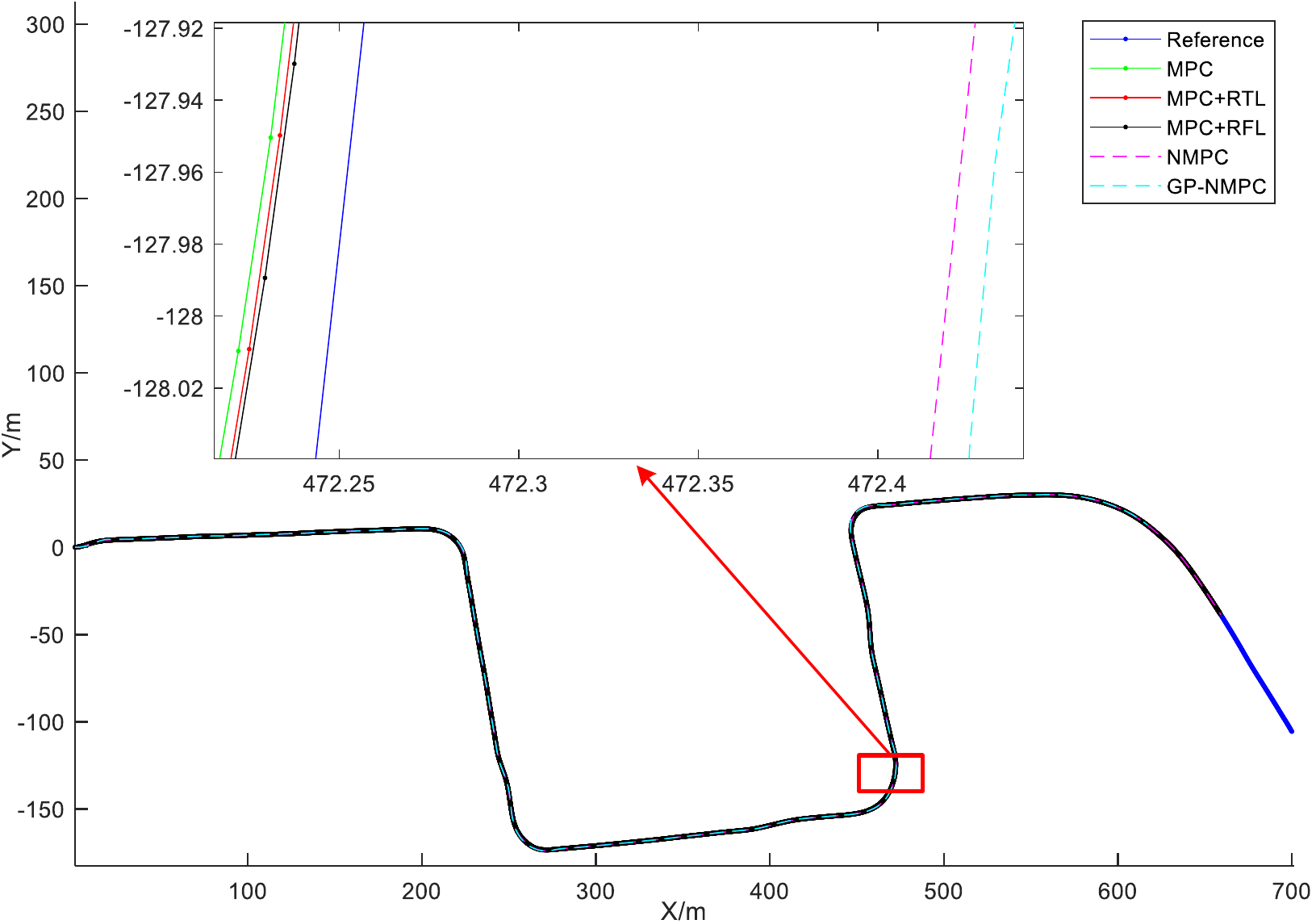}
		\end{minipage}
        \label{case2}
	}
	\caption{Residue prediction and fitting error with regression trees/ random forests and linear regression.}
	\label{result2}
\end{figure}

We also compared our tracking results with \cite{kabzan2019learning}, shown in TABLE \ref{table3} and Fig. \ref{result2}, and the correspondent code comes from a reproduced demo\footnote{https://github.com/lucasrm25/Gaussian-Process-based-Model-Predictive-Control} of  \cite{kabzan2019learning} with MATLAB. The MAE shows that the NMPC gets better than that of  MPC while MPC+RT/RF learning part can surpass the NMPC on the mean absolute lateral error. However, the GP-based NMPC  gets worse with the learning data of 300 points selected from enormous training data running by NMPC on the training path. A likely explanation for it is that the selection method suitable for the racing car with the purpose of minimum circle time is not suitable for the path following with the purpose of accurate tracking. To sum up, the MPC+RFL method gets the best accuracy on the test path tracking.

\begin{table}[h]
\caption{Quantitative Comparison of Lateral Error with Different Residual Models}
\label{table_example}
\begin{center}
\begin{tabularx}{0.48\textwidth}{X Y Y Y Y}
\toprule
Controller	
&MAE	&RMSE	&Max	&PE[\%]\\
\midrule
MPC	&0.0206	&0.0403	&0.2517	&-\\
MPC+RT	&0.0189	&0.0370	&0.2304	&7.89\\
MPC+RTL	&0.0176	&0.0344	&0.2118	&14.29\\
MPC+RF	&0.0187	&0.0366	&0.2280	&9.02\\
MPC+RFL	&\textbf{0.0172}&\textbf{0.0335}
&\textbf{0.2020}	&\textbf{16.32}\\
NMPC    &0.0200 &	0.0375 &	0.1786 &- \\
NMPC+GP	&0.2145 &	0.0759 &	0.3967 &- \\
\bottomrule
\end{tabularx}
\end{center}
\label{table3}
\end{table}

\begin{table}[h]
\caption{Computing Time Comparison with Different Residual Models}
\label{table_example}
\begin{center}
\begin{tabularx}{0.48\textwidth}{l Y Y Y Y c}
\toprule
&MPC	&+RTL	&+RFL	&NMPC	&NMPC+GP\\
\midrule
Mean time\ (s)	&0.059	&0.184	&0.695	&2.308	&15.434\\
Max time\ (s)	&0.827	&0.886	&1.641	&4.137	&21.692\\
Optimization\ (s)	&0.012	&0.027	&0.012	&1.676	&14.274\\
\bottomrule
\end{tabularx}
\end{center}
\label{table4}
\end{table}

\subsection{Computing Time Analysis}

 In the simulation experiment, the parameters are set as $Iter=30$, $m=300$, $F=8$, $D=6$, and $T=20$. Without multiple iterations of optimization, the pipeline we proposed could greatly save calculation time. The computing time on joint simulation of Matlab and Carsim is shown in TABLE \ref{table4}. The computing time of GP-NMPC is around 10 times MPC+RFL and 20 times MPC+RTL. With less computing time, the MPC+RTL/RFL pipeline could be widely used for more dynamic control problems. 

\section{Conclusion}

This paper has proposed a methodology to learn a switching residual model for MPC leveraging regression trees/random forests and linear approximation to enhance a simple error state nominal model and improve the tracking performance for autonomous driving. The residual model makes up for the mismatch of the nominal model and improves the prediction accuracy thus, improves the tracking accuracy. To obtain fast optimization results, the error state model and linear approximation are adopted, which keep the optimization of a QP problem. The simulation results show that this method could improve the tracking accuracy by about 16\%  while maintaining less time-consuming, and the mean time is 10 times less than the GP-NMPC method. Poor linear fitting problems on some leaves and other approximation methods will be studied in future work.

\addtolength{\textheight}{-12cm}   











\bibliographystyle{IEEEtran}
\bibliography{ref}

\end{document}